\documentclass[11pt]{article}

\usepackage{acl}

\usepackage{times}
\usepackage{latexsym}
\usepackage{booktabs}
\usepackage{multirow}
\usepackage{colortbl}
\usepackage{pgf}
\usepackage{amssymb}
\usepackage{tcolorbox}
\usepackage{todonotes}
\usepackage{graphicx}
\usepackage{subcaption}
\usepackage{longtable}
\usepackage{array}    
\usepackage{booktabs} 
\usepackage{numprint}
\npthousandsep{,}
\usepackage{colortbl}
\usepackage{nicematrix}
\usepackage{pgfplots}
\pgfplotsset{compat=1.18}
\usepgfplotslibrary{colormaps}
\usepackage{xfp}
\usepackage[table]{xcolor}
\usepackage{cleveref}

\usepackage{makecell}
\global

\tcbset{
  aibox/.style={
    width=474.18663pt,
    top=10pt,
    colback=white,
    colframe=black,
    colbacktitle=black,
    enhanced,
    center,
    attach boxed title to top left={yshift=-0.1in,xshift=0.15in},
    boxed title style={boxrule=0pt,colframe=white,},
  }
}
\newtcolorbox{AIbox}[2][]{aibox,title=#2,#1}

\definecolor{blue1}{RGB}{234, 230, 255} 
\definecolor{blue2}{RGB}{194, 200, 255} 
\definecolor{blue3}{RGB}{154, 170, 255}

\definecolor{red1}{RGB}{255,245,238}
\definecolor{red2}{RGB}{255,228,225}
\definecolor{red3}{RGB}{255,188,185}

\definecolor{light-gray}{HTML}{E5E4E2}
\definecolor{light-cyan}{HTML}{E0FFFF}
\newcolumntype{R}{>{\raggedleft\arraybackslash}p{0.28cm}}
\newcolumntype{L}{>{\raggedleft\arraybackslash}p{0.23cm}}

\newcommand\polardetect{\textcolor{black}{\textbf{\textsc{PolarDetect}}}}
\newcommand\polartype{\textcolor{black}{\textbf{\textsc{PolarType}}}}
\newcommand\polarmanifest{\textcolor{black}{\textbf{\textsc{PolarManifest}}}}
\newcommand\datasetname{\textcolor{black}{\textsc{POLAR}}}
\usepackage{microtype}
\usepackage{inconsolata}
\usepackage{graphicx}
\usepackage{enumitem}
\definecolor{lightblue}{RGB}{204, 229, 255}
\definecolor{lightorange}{RGB}{255, 229, 204}

\title{SemEval-2026 Task 9: Detecting Multilingual, Multicultural \\ and Multievent Online Polarization}

\author{
	Usman Naseem$^{1}$,
	Robert Geislinger$^{2}$,
	Juan Ren$^{1}$,
	Sarah Kohail$^{3}$,
	Rudy Garrido Veliz$^{2}$,
	\\ \bf
    P Sam Sahil$^{2,4}$, 
    Yiran Zhang$^{1}$, 
    Marco Antonio Stranisci$^{5,6}$,
    Idris Abdulmumin$^{7}$,
    \\ \bf
    Özge Alacam$^{8}$,
    Cengiz Acartürk$^{9}$, 
    Aisha Jabr$^{3}$, 
    Saba Anwar$^{2}$, 
    Abinew Ali Ayele$^{10}$, 
    \\ \bf
    Elena Tutubalina$^{11,12,13}$, 
    Aung Kyaw Htet$^{1}$, 
    Xintong Wang$^{2}$, 
    Surendrabikram Thapa$^{14}$, 
    \\ \bf 
    Tanmoy Chakraborty$^{15}$,
    Dheeraj Kodati$^{16}$, 
    Sahar Moradizeyveh$^{1}$,
    Firoj Alam$^{17,18}$, 
    \\ \bf
    Ye Kyaw Thu$^{19}$,
    Shantipriya Parida$^{20}$,
    Ihsan Ayyub Qazi$^{21}$,
    Lilian Wanzare$^{22}$,
	\\ \bf 
    Nelson Odhiambo Onyango$^{22}$,
	Clemencia Siro$^{23}$,
    Ibrahim Said Ahmad$^{24,25}$,
    \\ \bf 
    Adem Chanie Ali$^{2,10}$,
	Martin Semmann$^{2}$,
    Chris Biemann$^{2}$,
    \\ \bf 
	Shamsuddeen Hassan Muhammad$^{26}$,
	Seid Muhie Yimam$^{2}$
    \\[2mm]
	\footnotesize $^{1}$Macquarie University,
	\footnotesize $^{2}$University of Hamburg,
	\footnotesize $^{3}$Zayed University,
    \footnotesize $^{4}$HKBK College of Engineering,
    \footnotesize $^{5}$University of Turin,
	\\
    \footnotesize $^{6}$aequa-tech,
	\footnotesize $^{7}$University of Pretoria,
    \footnotesize $^{8}$Bielefeld University,
	\footnotesize $^{9}$Jagiellonian University,
	\footnotesize $^{10}$Bahir Dar University,
    \footnotesize $^{11}$AIRI,
    \\
    \footnotesize $^{12}$KFU,
    \footnotesize $^{13}$HSE University,
    \footnotesize $^{14}$Virginia Tech,
    \footnotesize $^{15}$IIT Delhi,
    \footnotesize $^{16}$ABV-IIITM,
    \footnotesize $^{17}$Qatar Computing Research Institute, 
    \\
    \footnotesize $^{18}$Hamad Bin Khalifa University,
    \footnotesize $^{19}$Language Understanding Lab., Myanmar,
    \footnotesize $^{20}$AMD Silo AI,
    \\
    \footnotesize $^{21}$Lahore University of Management Sciences,
    \footnotesize $^{22}$Maseno University,
    \footnotesize $^{23}$Centrum Wiskunde \& Informatica,
    \\
	\footnotesize $^{24}$Bayero University Kano,
	\footnotesize $^{25}$Northeastern University,
	\footnotesize $^{26}$Imperial College London,
    \\[1mm]
    \footnotesize \texttt{Contact: usman.naseem@mq.edu.au and  seid.muhie.yimam@uni-hamburg.de}
}

\begin{document}
\maketitle
\begin{abstract}

We present SemEval-2026 Task 9, a shared task on online polarization detection, covering 22 languages and comprising over 110K annotated instances.
Each data instance is multi-labeled with the presence of polarization, polarization type, and polarization manifestation.
Participants were asked to predict labels in three subtasks: (1) detecting the presence of polarization, (2) identifying the type of polarization, and (3) recognizing the polarization manifestation.
The three tasks attracted over 1,000 participants worldwide and more than 10k submissions on Codabench.
We received final submissions from 67 teams and 69 system description papers.
We report the baseline results and analyze the performance of the best-performing systems, highlighting the most common approaches and the most effective methods across different subtasks and languages.
The dataset and other resources for this task are publicly available.\footnote{\url{https://polar-semeval.github.io/}}

\end{abstract}

\section{Introduction}

Online polarization, defined as sharp division and antagonism between social, political, or identity groups, has become a pervasive threat to democratic institutions, civil discourse, and social cohesion worldwide~\citep{waller2021quantifying}.
It is often fueled by biased or inflammatory content in digital media, strengthening echo chambers and undermining mutual understanding~\citep{garimella2018polarization}.
Polarized discourse amplifies ideological divides and can escalate into hate speech, harassment, and real-world violence \cite{piazza2023political,martinez2024methodology}.
Therefore, early detection of polarization is essential for designing interventions that promote healthier online ecosystems.

\begin{figure}
    \centering
    \includegraphics[width=1\linewidth]{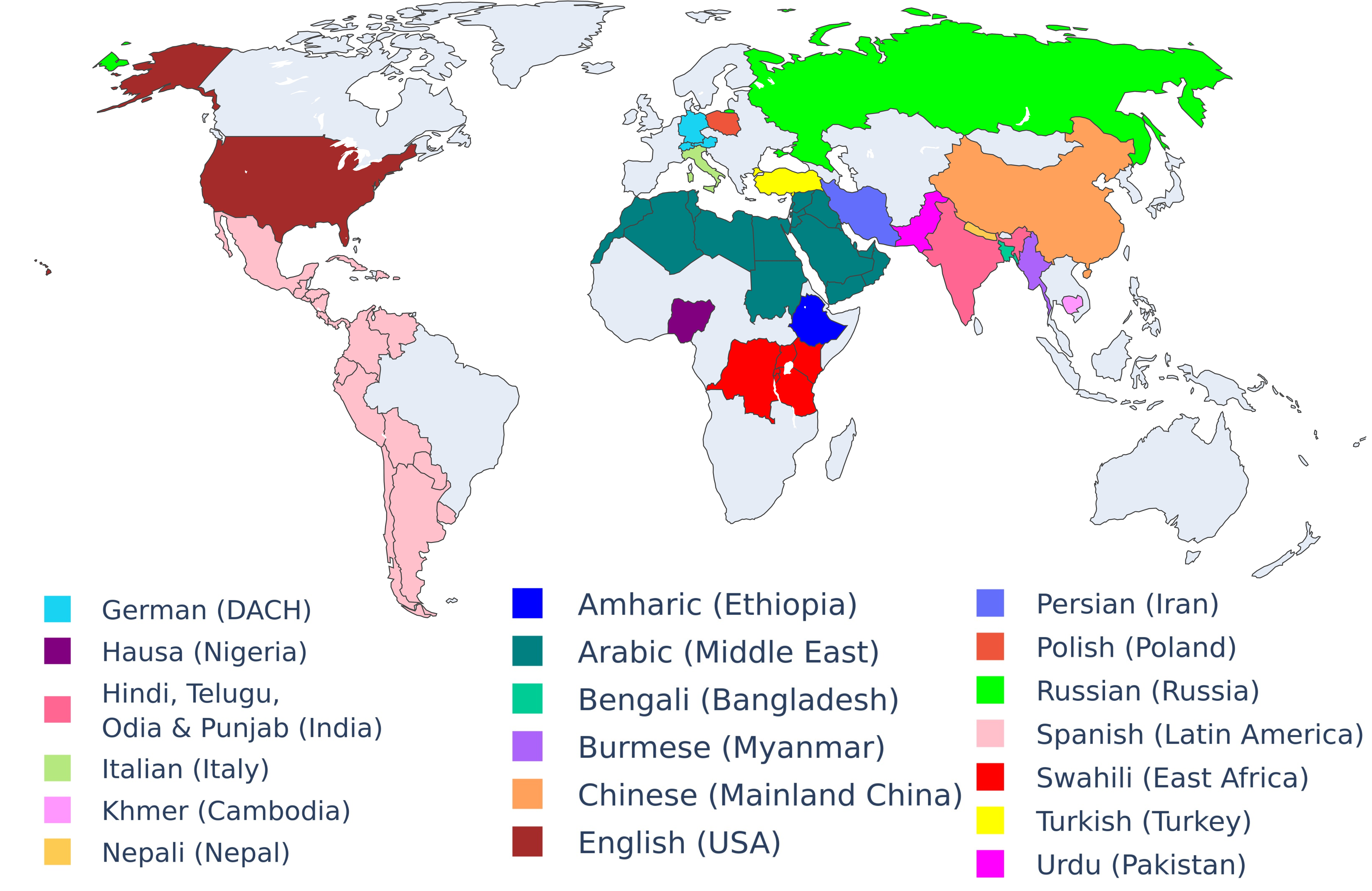}
    \caption{Languages represented in a world map covered by \datasetname, covering diverse linguistic and regional contexts. The language and societal context can present themselves across varied areas. Language assignments to countries and regions are approximate.}
    \label{fig:lang-map}

\end{figure}

In this shared task, we provide participants with 
\datasetname{}, a large-scale, multilingual, 
multicultural, and multi-event dataset for fine-grained 
polarization detection~\cite{naseem2026polarbenchmarkmultilingualmulticultural}. 
The task challenges participants to develop systems 
that can automatically detect and classify polarized 
content across multiple languages, cultural contexts, 
and event types. \datasetname{} covers 22 languages 
spanning seven language families and comprises over
110,000 annotated instances (see Figure~\ref{fig:lang-map} for the geographic and linguistic diversity represented). Table~\ref{tab:data_distribution_IAA} presents the data distribution across the train, development, and test splits. This shared task supports three complementary subtasks:

\begin{itemize}[noitemsep,leftmargin=*]
   
\item \textbf{Binary Polarization Detection:}  Determine whether a given text expresses polarization. We refer to this task as  \polardetect. 

 \item \textbf{Polarization Type Classification:} Identify the social dimension underlying polarization
 (e.g., political, religious, racial). We refer to this task as \polartype. 

\item \textbf{Manifestation Identification:} Detect how polarization is rhetorically manifested, including strategies such as stereotyping, deindividuation, vilification, dehumanization, extreme language, and other rhetorical devices. We refer to this task as \polarmanifest. 

Each team could submit results for subtask 1, 2, 3, or all three subtasks in one or more languages. Our official evaluation metrics were the average Macro F~1. Our tasks attracted over 1000 participants, with 548 final submissions in the test phase and 69 system description papers. Subtask~1 received the most submissions (267), followed by subtask~2 with 161, and subtask~3 with 120.

\end{itemize}

\begin{table}[!t]
\centering
\scriptsize
\setlength{\tabcolsep}{4pt}
\renewcommand{\arraystretch}{1}
\begin{tabular}{l|ccc|c|c}
\toprule
\textbf{Lang.} &\textbf{ Train} & \textbf{Dev }& \textbf{Test} & \textbf{Total} & \textbf{Inner Agr.} ($\kappa$) \\
\midrule
amh & 3,332 & 166 & 1,501 & 4,999 & 0.59 \\
arb & 3,380 & 169 & 1,521 & 5,070 & 0.25 \\
ben & 3,333 & 166 & 1,501 & 5,000 & 0.59 \\
deu & 3,180 & 159 & 1,432 & 4,771 & 0.10* \\
eng & 3,222 & 160 & 1,452 & 4,834 & 0.39 \\
fas & 3,295 & 164 & 1,484 & 4,943 & 0.78 \\
hau & 3,651 & 182 & 1,644 & 5,477 & 0.48 \\
hin & 2,744 & 137 & 1,236 & 4,117 & 0.49 \\
ita & 3,334 & 166 & 1,538 & 5,038 & 0.39 \\
khm & 6,640 & 332 & 2,988 & 9,960 & 0.83 \\
mya & 2,889 & 144 & 1,301 & 4,334 & 0.13 \\
nep & 2,005 & 100 & 903 & 3,008 & 0.79 \\
ori & 2,368 & 118 & 1,066 & 3,552 & 0.46 \\
pan & 1,700 & 100 & 809 & 2,609 & 0.55* \\
pol & 2,391 & 119 & 1,077 & 3,587 & 0.46 \\
rus & 3,348 & 167 & 1,508 & 5,023 & 0.39 \\
spa & 3,305 & 165 & 1,488 & 4,958 & 0.26 \\
swa & 6,991 & 349 & 3,147 & 10,487 & 0.56 \\
tel & 2,366 & 118 & 1,066 & 3,550 & 0.7 \\
tur & 2,364 & 115 & 1,093 & 3,572 & 0.46 \\
urd & 3,563 & 177 & 1,606 & 5,346 & 0.29 / 0.70* \\
zho & 4,280 & 214 & 1,927 & 6,421 & 0.64 \\
\midrule
\textbf{Total} & \textbf{73,681} & \textbf{3,687} & \textbf{33,288} & \textbf{110,656} & \\
\bottomrule
\end{tabular}
\caption{Data distribution across the train, development, and test splits, along with inner agreement. Inner Agre. denotes inter-annotator agreement per language (Fleiss’s $\kappa$ unless otherwise noted). $^{*}$ denotes exceptions: German uses Krippendorff’s $\alpha$; Punjabi reports identical Krippendorff’s $\alpha$ and Cohen’s $\kappa$; Urdu reports Fleiss’s $\kappa$ / Cohen’s $\kappa$. }
\label{tab:data_distribution_IAA}
\end{table}

\section{Related Work}

Online polarization poses a threat to social cohesion, exacerbated by social media echo chambers and biased content~\citep{waller2021quantifying, iandoli2021impact, garimella2018polarization}. As social media and other online platforms become key arenas for political and cultural discourse, the need for early detection and nuanced understanding of polarization has grown significantly. Polarization detection is important for content moderation, peace building, responsible digital governance, and healthy democracy. Foundational research has defined polarization as both intergroup hostility and blind ingroup cohesion~\citep{arora2022polarization}, and has highlighted its relationship with hate speech, fragmentation, and incivility~\citep{ mathew2020hatexplain}.

A growing body of research has documented the role of online spaces in intensifying polarization across different regions \citep{kubin2021role, barbera2020social, gitlin2016outrage, soares2021hashtag}. However, most computational work focuses on high-resource languages and event- or region-specific datasets, limiting generalizability~\citep{kubin2021role}. This leaves a significant gap in our ability to generalize findings across cultures, languages, and events, especially for low-resource languages or multilingual regions.

The lack of standardized datasets across languages has hindered progress in developing and evaluating polarization detection models with cross-lingual or cross-cultural capabilities. Recent shared tasks on hate speech and toxicity~\citep{basile2019semeval, pamungkas2020misogynistic} have expanded the language and domain coverage, yet remain less fine-grained regarding polarization’s diverse types and rhetorical manifestations. This shared task addresses this gap by presenting a comprehensive, fine-grained dataset benchmark for multilingual, multicultural, and multievent online polarization, enabling robust cross-lingual and context-aware modeling.

\section{\datasetname~Dataset Construction}

\subsection{Operational Definitions}
\label{polardef}

Our work~\cite{naseem2026polarbenchmarkmultilingualmulticultural} defines polarization as the increasing extremity of opinions, beliefs, or behaviors, resulting in heightened inter-group divisions and conflict. Besides, we defined polarization types including political, racial or ethnic, religious, gender or sexual, ant other. We further distinguish polarization by its rhetorical manifestations, containing stereotype, vilification, dehumanization, extreme language, lack of Empathy, and Invalidation

\subsection{Data Collection}
We collected data from a range of online platforms, including major social media sites, local news, and commentary forums.
For several languages, including Burmese, Polish, and Chinese, we sampled and re-annotated instances from existing toxic or hate speech datasets.

The curated dataset covers diverse real-world events, grounding event selection in the sociopolitical and socioeconomic contexts specific to each language and cultural setting.
The data span a broad range of events and issues, including armed conflicts, elections and party politics, public health crises, large-scale migration, climate change, and broader socioeconomic debates. The dataset also includes discussions related to gender and indigenous rights, religion, and ideology.

We provide more detailed information about the definitions of the categories, annotation guidelines, collected events, and data processing in detail in \citet{naseem2026polarbenchmarkmultilingualmulticultural}.

\subsection{Annotation Process and Guidelines}

We used a hybrid annotation strategy, leveraging crowd-sourced annotators and trained community annotators for low-resource languages where crowd-sourced annotation support is limited.
For the crowd-sourced setting, we used Mechanical Turk\footnote{\url{https://www.mturk.com}} and Prolific\footnote{\url{https://www.prolific.com}}, and annotators were selected based on their prior experience and annotation quality.
Specifically, we filtered candidates using historical annotation agreement scores and conducted pilot rounds to identify those with consistent performance.

Given the cultural and linguistic breadth of \datasetname, we developed detailed, multilingual annotation guidelines in English, and then translated and culturally adapted them for each target language.

\begin{tcolorbox}
Annotators were instructed to:
\begin{itemize}[noitemsep,leftmargin=*]
 \item Identify whether a text is polarized
 \item If the text is classified as polarized, tag the type of polarization (political, racial/ethnic, religious, gender/sexual identity, other)
 \item If the text is classified as polarized, tag its manifestations or rhetorical tactics (stereotyping/deindividuation, vilification, dehumanization, extreme language, lack of empathy, invalidation).
\end{itemize}
\end{tcolorbox}


Multiple labels were allowed due to the conceptual and contextual overlap often observed in polarized content.
The details about the guidelines, annotation process, and annotator reliability are described in \citep{naseem2026polarbenchmarkmultilingualmulticultural}.

\subsection{Annotators’ Reliability} 
To evaluate annotation quality, we report Fleiss’ Kappa, Cohen’s kappa, and Krippendorff’s alpha as inter-annotator agreement (IAA) metrics. Different metrics are used because the annotation setups differ across languages, having different numbers of annotators per instance~\cite{10.1162/coli.07-034-R2}. As shown in \Cref{tab:data_distribution_IAA}, the IAA scores vary between languages, with the majority showing moderate agreement and a few, such as ``khm'' and ``tel'' achieving good agreement. 
Although guidelines were standardized, their interpretation was influenced by cultural and political context, especially in languages with lower agreement, where some terms may not have direct equivalents across cultures. 
 Latent content or sarcasm often required annotators to draw on their own socio-political knowledge, highlighting the perspectivist nature of polarization \citep{cabitza2023toward}. Thus, low agreement can indicate socio-pragmatic complexity rather than error, signaling that polarization markers may not have universal meanings and that divergences can reveal inherent ambiguity in stimuli or interpretation \citep{aroyo2015truth}.
 
\section{Task Description}

The participants received the data of texts from different sources and different lengths.
They were instructed to classify the texts on polarization and its components. The task comprised three subtasks, of which the participants could choose to participate in one or more.
\subsection{Subtasks}
\paragraph{Subtask 1: \polardetect}
The participants had to correctly assign whether the text was polarized or not polarized, a straightforward binary decision based on the definition of polarization used. All 22 languages were available in this subtask.

\paragraph{Subtask 2: \polartype}
Based on a polarized text selected in \polardetect, the participants were asked to assign the text into a type of polarization: political, racial or ethnic, religious, gender or sexual, or other (based on economic class, media, etc.). 
All 22 languages were available in this subtask as well.

\paragraph{Subtask 3: \polarmanifest}
Given the polarized text, and the type(s) of polarization of that text (i.e., political, racial or ethnic, religious, gender or sexual, or other), participants had to correctly predict the label of manifestation(s) of the polarized text: stereotype, vilification, dehumanization, extreme language and absolutism, lack of empathy, or invalidation. The languages: Burmese (mya), Italian (ita), Polish (pol), and Russian (rus) were not present in this subtask. Resulting in the data available for 18 languages. 

\subsection{Task Organisation}

We used Codabench as the competition platform, setting up three different competitions, one for each subtask, to allow individual participation.

\begin{figure}[!t]
    \centering
    \includegraphics[width=1\linewidth]{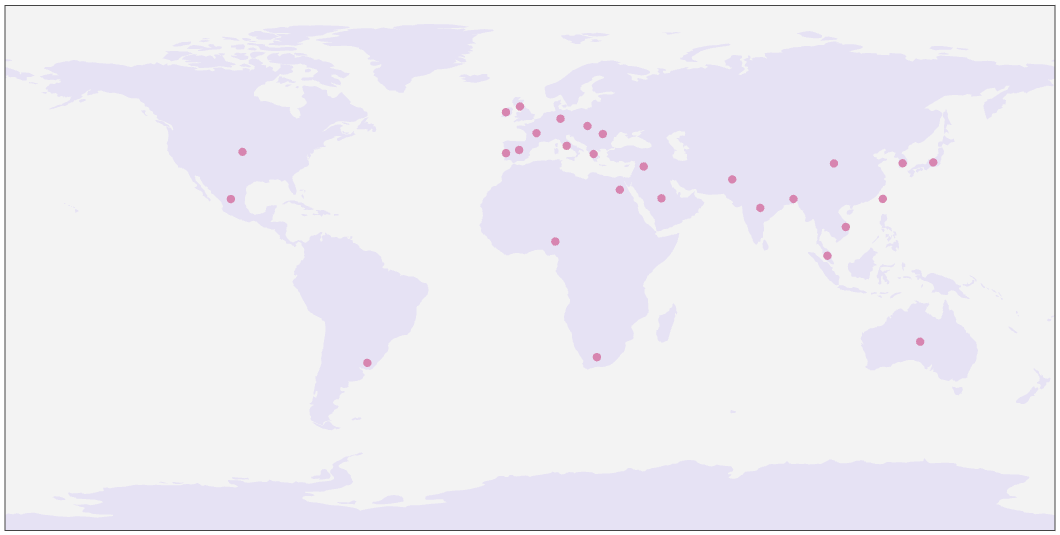}
    \caption{Participants came from 28 unique countries and regions: Australia, Bangladesh, China, Egypt, France, Germany, Greece, India, Ireland, Italy, Japan, Malaysia, Mexico, Nigeria, Pakistan, Portugal, Romania, Saudi Arabia, Slovakia, South Africa, South Korea, Spain, Syria, Taiwan, United Kingdom, United States, Uruguay, and Vietnam.}
    \label{fig:participants_map}

\end{figure}

We released pilot datasets before the start of the shared task to help participants better understand the task, such as data structure, the language involved, and the labels.
We provided participants with a starter kit on GitHub, resources for beginners, and organized a Q\&A session along with a writing tutorial for junior researchers.
Participants were also supported with more details on each task, and their concerns were answered throughout the Discord server of the task and through emails forwarded to organizers.
Our participants were based in different parts of the world, as shown in Figure~\ref{fig:participants_map}.
The task consisted of two phases: (1) the development phase and (2) the evaluation phase.
During the development phase, the leaderboard was open, allowing a maximum of 999 submissions per participant.
In the evaluation phase, the leaderboard was closed, and each participant was allowed up to five submissions.
Only the last submission is being considered for the official ranking.

\subsection{Evaluation Metrics and Baselines}

\paragraph{Evaluation Metrics} We use the average macro F1\- score for participants' results by comparing predicted and the gold-standard labels.

\paragraph{Our Baselines} We provide our baseline for each language by applying LaBSE~\cite{liu2019robertarobustlyoptimizedbert}. We finetuned LaBSE using the training data for each language for all three subtasks. Table~ \ref{tab:subtask1}, \ref{tab:subtask2}, and \ref{tab:subtask3} show the average macro F1 of top-performing systems compared to our baseline in all three subtasks.

\section{Participating Systems and Results}

POLAR was the most popular SemEval competition on Codabench in 2026. Our three subtasks rank 1st, 3rd, and 7th in popularity among the 18 subtasks across the 12 shared tasks in SemEval 2026\footnote{\url{https://www.codabench.org/competitions/public}}. Our shared task attracted over 1,000 participants from 28 countries and regions worldwide, as illustrated in Figure~\ref{fig:participants_map}. Specifically, POLAR attracted 533 participants in Subtask~1, 344 in Subtask~2, and 185 participants in Subtask~3 (see Table~\ref{tab:task_participants_stats}).
In the development phase, more than 5.7K submissions were made to Subtask~1, more than 2.5K to Subtask~2, and over 1k to Subtask~3. In the test phase, 267 submissions were made to Subtask~1, 161 for Subtask~2, and 120 for Subtask~3. The official results included 67 teams and 69 system description papers; one team submitted three papers, one for each subtask. Overall, 43\% of teams participated in only one subtask, 16\% in two subtasks, and 41\% in all three subtasks. Participants generally preferred to submit systems for all languages rather than a subset of languages. Specifically, 41\% of teams in subtask~1 submitted predictions for all languages, compared to 56\% and 63\% in subtask~2 and subtask~3.

\begin{table}[!h]
\centering
\scriptsize
\setlength{\tabcolsep}{2.5pt}
\resizebox{\columnwidth}{!}{
\begin{tabular}{lcccc}
\toprule
\multirow{2}{*}{\textbf{Subtask}} & 
\multirow{2}{*}{\textbf{Participants}} & 
\multirow{2}{*}{\shortstack{\textbf{Dev}\\\textbf{Submissions}}} & 
\multirow{2}{*}{\shortstack{\textbf{Test}\\\textbf{Submissions}}} & 
\multirow{2}{*}{\shortstack{\textbf{Teams} \\\textbf{in Results}}} \\
 & & & & \\
\midrule
1 & 533 & 5,764 & 267 & 56 \\
2 & 344 & 2,555 & 161 & 39 \\
3 & 185 & 1,029 & 110 & 25 \\
\midrule
\textbf{Total} & \textbf{1,061} & \textbf{9,886} &\textbf{548} & \textbf{69} \\
\bottomrule
\end{tabular}
}
\caption{Participation statistics for the POLAR shared task on Codabench.``Teams in Results'' are those that submitted system description papers. In total, 67 teams with 69 papers appear on the leaderboard across all subtasks. }
\label{tab:task_participants_stats}
\end{table}

\begin{table*}[!t]
\centering
\scriptsize
\setlength{\tabcolsep}{2.5pt}
\renewcommand{\arraystretch}{0.9}
\resizebox{\textwidth}{!}{%
\begin{tabular}{lcccc@{\hspace{0.6cm}}lcccc@{\hspace{0.6cm}}lcccc}
\cmidrule(r){1-5} \cmidrule(r){6-10} \cmidrule(r){11-15}
\multicolumn{5}{c}{\textbf{Subtask 1}} & \multicolumn{5}{c}{\textbf{Subtask 2}} & \multicolumn{5}{c}{\textbf{Subtask 3}} \\
\cmidrule(r){1-5} \cmidrule(r){6-10} \cmidrule(r){11-15}
\textbf{Team} & \textbf{Total} & \textbf{1st} & \textbf{2nd} & \textbf{3rd} &
\textbf{Team} & \textbf{Total} & \textbf{1st} & \textbf{2nd} & \textbf{3rd} &
\textbf{Team} & \textbf{Total} & \textbf{1st} & \textbf{2nd} & \textbf{3rd} \\
\cmidrule(r){1-5} \cmidrule(r){6-10} \cmidrule(r){11-15}
\cellcolor{blue3}UTokyo Tsuruoka Lab & 12 & 8 & 4 & 0 & \cellcolor{blue3}UTokyo Tsuruoka Lab & 13 & 7 & 5 & 1 & \cellcolor{blue3}SMASH & \multicolumn{1}{r}{16} & \multicolumn{1}{r}{9} & \multicolumn{1}{r}{4} & \multicolumn{1}{r}{3} \\
\cellcolor{blue2}NYCU-NLP & 12 & 3 & 5 & 4 & \cellcolor{blue2}NYCU-NLP & 15 & 6 & 5 & 4 & \cellcolor{blue2}NYCU-NLP & \multicolumn{1}{r}{11} & \multicolumn{1}{r}{7} & \multicolumn{1}{r}{3} & \multicolumn{1}{r}{1} \\
\cellcolor{blue1}PSK & 9 & 2 & 4 & 3 & \cellcolor{blue1}SMASH & 13 & 4 & 6 & 3 & \cellcolor{blue1}Sagarmatha & \multicolumn{1}{r}{4} & \multicolumn{1}{r}{2} & \multicolumn{1}{r}{0} & \multicolumn{1}{r}{2} \\
CYUT & 4 & 2 & 0 & 2 & Lingo Research Group & 7 & 2 & 1 & 4 & Ping An & \multicolumn{1}{r}{4} & \multicolumn{1}{r}{0} & \multicolumn{1}{r}{4} & \multicolumn{1}{r}{0} \\
SMASH & 7 & 1 & 2 & 4 & PolaFusion & 4 & 1 & 0 & 3 & PolaFusion & \multicolumn{1}{r}{7} & \multicolumn{1}{r}{0} & \multicolumn{1}{r}{2} & \multicolumn{1}{r}{5} \\
Lingo Research Group & 5 & 1 & 1 & 3 & Sagarmatha & 2 & 1 & 0 & 1 & OZemi & \multicolumn{1}{r}{3} & \multicolumn{1}{r}{0} & \multicolumn{1}{r}{2} & \multicolumn{1}{r}{1} \\
taien & 3 & 1 & 1 & 1 & AIvengers & 1 & 0 & 1 & 0 & AIvengers & \multicolumn{1}{r}{4} & \multicolumn{1}{r}{0} & \multicolumn{1}{r}{1} & \multicolumn{1}{r}{3} \\
OZemi & 2 & 1 & 0 & 1 & ShefFriday & 1 & 0 & 1 & 0 & CYUT & \multicolumn{1}{r}{1} & \multicolumn{1}{r}{0} & \multicolumn{1}{r}{1} & \multicolumn{1}{r}{0} \\
Sagarmatha & 1 & 1 & 0 & 0 & Stochastic Gradient Descenders & 1 & 0 & 1 & 0 & ShefFriday & \multicolumn{1}{r}{1} & \multicolumn{1}{r}{0} & \multicolumn{1}{r}{1} & \multicolumn{1}{r}{0} \\
mdok-style & 1 & 1 & 0 & 0 & MSqrd & 1 & 0 & 1 & 0 & YEZE & \multicolumn{1}{r}{2} & \multicolumn{1}{r}{0} & \multicolumn{1}{r}{0} & \multicolumn{1}{r}{2} \\
PhatThachDau & 1 & 1 & 0 & 0 & CYUT & 1 & 0 & 0 & 1 & Lingo Research Group & \multicolumn{1}{r}{1} & \multicolumn{1}{r}{0} & \multicolumn{1}{r}{0} & \multicolumn{1}{r}{1} \\
MKJ & 2 & 0 & 2 & 0 & YEZE & 1 & 0 & 0 & 1 &  &  &  &  &  \\
StanceLab & 2 & 0 & 2 & 0 & mdok-style & 1 & 0 & 0 & 1 &  &  &  &  &  \\
CUET-823 & 1 & 0 & 1 & 0 & YNU-HPCC & 1 & 0 & 0 & 1 &  &  &  &  &  \\
PolDeck & 1 & 0 & 1 & 0 & PolarMind & 1 & 0 & 0 & 1 &  &  &  &  &  \\
Projet Fil Rouge 821 & 1 & 0 & 1 & 0 &  & \multicolumn{1}{l}{} & \multicolumn{1}{l}{} & \multicolumn{1}{l}{} & \multicolumn{1}{l}{} &  &  &  &  &  \\
UMUSP & 1 & 0 & 1 & 0 &  & \multicolumn{1}{l}{} & \multicolumn{1}{l}{} & \multicolumn{1}{l}{} & \multicolumn{1}{l}{} &  &  &  &  &  \\
PolaFusion & 1 & 0 & 0 & 1 &  & \multicolumn{1}{l}{} & \multicolumn{1}{l}{} & \multicolumn{1}{l}{} & \multicolumn{1}{l}{} &  &  &  &  &  \\
YEZE & 1 & 0 & 0 & 1 &  & \multicolumn{1}{l}{} & \multicolumn{1}{l}{} & \multicolumn{1}{l}{} & \multicolumn{1}{l}{} &  &  &  &  &  \\
MoMo & 1 & 0 & 0 & 1 &  & \multicolumn{1}{l}{} & \multicolumn{1}{l}{} & \multicolumn{1}{l}{} & \multicolumn{1}{l}{} &  &  &  &  &  \\
Semantic Vectors & 1 & 0 & 0 & 1 &  & \multicolumn{1}{l}{} & \multicolumn{1}{l}{} & \multicolumn{1}{l}{} & \multicolumn{1}{l}{} &  &  &  &  &  \\
Tralaleros & 1 & 0 & 0 & 1 &  & \multicolumn{1}{l}{} & \multicolumn{1}{l}{} & \multicolumn{1}{l}{} & \multicolumn{1}{l}{} &  &  &  &  & \\

\cmidrule(r){1-5} \cmidrule(r){6-10} \cmidrule(r){11-15}
\end{tabular}%
}
\caption{Top-3 placements achieved by teams across the three subtasks. For each task, the table reports the total number of top-3 finishes achieved by each team and their breakdown into \colorbox{blue3}{1st}, \colorbox{blue2}{2nd}, and \colorbox{blue1}{3rd} places.}
\label{tab:medal_by_task}
\end{table*}

We report results only for teams that submitted a system description paper.
Table~\ref{tab:medal_by_task} summarizes the distribution of top-3 placements across subtasks.
Table~\ref{tab:subtask1}
presents the results for Subtask 1, which had 79 participating teams. Table~\ref{tab:subtask2}
shows the results for Subtask 2, with
47 participating teams, while Table~\ref{tab:subtask3}
reports the results for Subtask 3, which had 30 participating teams.

\subsection{Subtask 1: \polardetect}

\begin{table*}[t]
\centering
\setlength{\tabcolsep}{2.5pt}
\renewcommand{\arraystretch}{0.9}
\scriptsize
\resizebox{\textwidth}{!}{%
\begin{tabular}{lll|lll|lll|lll}
\toprule
\textbf{Lang} & \textbf{Team} & \textbf{Score} &
\textbf{Lang} & \textbf{Team} & \textbf{Score} &
\textbf{Lang} & \textbf{Team} & \textbf{Score} &
\textbf{Lang} & \textbf{Team} & \textbf{Score} \\
\midrule

\multirow{4}{*}{amh} & \cellcolor{blue1}PSK & \cellcolor{blue1}0.800 & \multirow{4}{*}{hau} & \cellcolor{blue1}PhatThachDau & \cellcolor{blue1}0.834 & \multirow{4}{*}{pan} & \cellcolor{blue1}UTokyo Tsuruoka Lab & \cellcolor{blue1}0.826 & \multirow{4}{*}{rus} & \cellcolor{blue1}UTokyo Tsuruoka Lab & \cellcolor{blue1}0.830 \\
 & UTokyo Tsuruoka Lab & 0.795 &  & Projet Fil Rouge 821 & 0.832 &  & PSK & 0.812 &  & NYCU-NLP & 0.823 \\
 & Lingo Research Group & 0.793 &  & OZemi & 0.831 &  & NYCU-NLP & 0.811 &  & CYUT & 0.814 \\
 & \cellcolor{red1}\textit{baseline} & \cellcolor{red1}0.764 &  & \cellcolor{red1}\textit{baseline} & \cellcolor{red1}0.821 &  & \cellcolor{red1}\textit{baseline} & \cellcolor{red1}0.749 &  & \cellcolor{red1}\textit{baseline} & \cellcolor{red1}0.748 \\
\midrule
\multirow{4}{*}{arb} & \cellcolor{blue1}UTokyo Tsuruoka Lab & \cellcolor{blue1}0.849 & \multirow{4}{*}{hin} & \cellcolor{blue1}CYUT & \cellcolor{blue1}0.828 & \multirow{4}{*}{tel} & \cellcolor{blue1}Sagarmatha & \cellcolor{blue1}0.905 & \multirow{4}{*}{spa} & \cellcolor{blue1}UTokyo Tsuruoka Lab & \cellcolor{blue1}0.803 \\
 & PSK & 0.848 &  & PSK & 0.824 &  & SMASH & 0.901 &  & NYCU-NLP & 0.800 \\
 & NYCU-NLP & 0.843 &  & Lingo Research Group & 0.821 &  & Tralaleros & 0.897 &  & SMASH & 0.798 \\
 & \cellcolor{red1}\textit{baseline} & \cellcolor{red1}0.812 &  & \cellcolor{red1}\textit{baseline} & \cellcolor{red1}0.782 &  & \cellcolor{red1}\textit{baseline} & \cellcolor{red1}0.889 &  & \cellcolor{red1}\textit{baseline} & \cellcolor{red1}0.750 \\
\midrule
\multirow{4}{*}{ben} & \cellcolor{blue1}UTokyo Tsuruoka Lab & \cellcolor{blue1}0.863 & \multirow{4}{*}{khm} & \cellcolor{blue1}SMASH & \cellcolor{blue1}0.774 & \multirow{4}{*}{tur} & \cellcolor{blue1}NYCU-NLP & \cellcolor{blue1}0.833 & \multirow{4}{*}{swa} & \cellcolor{blue1}PSK & \cellcolor{blue1}0.811 \\
 & CUET-823 & 0.858 &  & StanceLab & 0.761 &  & UTokyo Tsuruoka Lab & 0.830 &  & SMASH & 0.810 \\
 & NYCU-NLP & 0.854 &  & Semantic Vectors & 0.755 &  & PSK & 0.809 &  & taien & 0.799 \\
 & \cellcolor{red1}\textit{baseline} &\cellcolor{red1}0.825 &  & \cellcolor{red1}\textit{baseline} & \cellcolor{red1}0.737 &  \cellcolor{red1}& \textit{baseline}& \cellcolor{red1}0.750 &  & \cellcolor{red1}\textit{baseline} & \cellcolor{red1}0.790 \\
\midrule
\multirow{5}{*}{ita} & \cellcolor{blue1}mdok-style & \cellcolor{blue1}0.730 & \multirow{5}{*}{fas} & \cellcolor{blue1}OZemi & \cellcolor{blue1}0.835 & \multirow{5}{*}{mya} & \cellcolor{blue1}taien & \cellcolor{blue1}0.891 & \multirow{5}{*}{pol} & \cellcolor{blue1}Lingo Research Group & \cellcolor{blue1}0.843 \\
 & StanceLab & 0.672 &  & taien & 0.831 &  & MKJ & 0.887 &  & NYCU-NLP & 0.835 \\
 & PolaFusion & 0.671 &  & MKJ & 0.831 &  & NYCU-NLP & 0.887 &  & PSK & 0.835 \\
 & MoMo & 0.671 &  & PSK & 0.828 &  & SMASH & 0.885 &  & SMASH & 0.828 \\
 & \cellcolor{red1}\textit{baseline} & \cellcolor{red1}0.564 &  & \cellcolor{red1}\textit{baseline} & \cellcolor{red1}0.835 &  & \cellcolor{red1}\textit{baseline} & \cellcolor{red1}0.861 &  & \cellcolor{red1}\textit{baseline} & \cellcolor{red1}0.773 \\
\midrule
\multirow{4}{*}{deu} & \cellcolor{blue1}NYCU-NLP & \cellcolor{blue1}0.761 & \multirow{4}{*}{nep} & \cellcolor{blue1}NYCU-NLP & \cellcolor{blue1}0.924 & \multirow{4}{*}{urd} & \cellcolor{blue1}UTokyo Tsuruoka Lab & \cellcolor{blue1}0.820 &  &  & \multicolumn{1}{l}{} \\
 & UTokyo Tsuruoka Lab & 0.753 &  & Lingo Research Group & 0.918 &  & NYCU-NLP & 0.817 &  &  & \multicolumn{1}{l}{} \\
 & CYUT & 0.747 &  & SMASH & 0.914 &  & Lingo Research Group & 0.816 &  &  & \multicolumn{1}{l}{} \\
 & \textit{baseline} & 0.686 &  & \textit{baseline} & 0.883 &  & \textit{baseline} & 0.742 &  &  & \multicolumn{1}{l}{} \\
\cmidrule{1-9}
\multirow{4}{*}{eng} & \cellcolor{blue1}UTokyo Tsuruoka Lab & \cellcolor{blue1}0.825 & \multirow{4}{*}{ori} & \cellcolor{blue1}UTokyo Tsuruoka Lab & \cellcolor{blue1}0.826 & \multirow{4}{*}{zho} & \cellcolor{blue1}CYUT & \cellcolor{blue1}0.932 &  &  & \multicolumn{1}{l}{} \\
 & PolDeck & 0.819 &  & UMUSP & 0.814 &  & UTokyo Tsuruoka Lab & 0.929 &  &  & \multicolumn{1}{l}{} \\
 & PSK & 0.818 &  & YEZE & 0.812 &  & NYCU-NLP & 0.927 &  &  & \multicolumn{1}{l}{} \\
 & \cellcolor{red1}\textit{baseline} & \cellcolor{red1}0.773 &  & \cellcolor{red1}\textit{baseline} & \cellcolor{red1}0.776 &  & \cellcolor{red1}\textit{baseline} & \cellcolor{red1}0.864 &  &  & \multicolumn{1}{l}{}\\
\cmidrule{1-9}
\end{tabular}%
}
\caption{Top three performing systems for each language in subtask~1 evaluated using macro-F1 score.}
\label{tab:subtask1}
\end{table*}

\subsubsection{Best Performing Systems}

\textbf{Team UTokyo Tsuruoka Lab} achieved one of the strongest performances in the competition, ranking first in 8 out of 22 languages. Their system is based on the instruction-tuned Gemma-3-12B-IT \cite{gemmateam2025gemma3technicalreport} model and introduces an efficient one-forward-pass strategy for both training and inference. To enable memory-efficient fine-tuning of the large language model, they utilized Unsloth \citep{unsloth}, which reduces GPU memory requirements during training. A key aspect of their approach is a selective-token training method, where the model predicts labels through one-token inference rather than using a traditional multi-label classification head. This formulation simplifies the prediction process and improves inference efficiency \cite{semeval2026_task9_utokyo_tsuruoka_lab}.

\textbf{Team NYCU-NLP} proposed a system based on instruction-tuned small language models, including Gemma-3 (27B) ~\cite{gemmateam2025gemma3technicalreport}, Mistral Small 3.2 (24B) ~\cite{mistral_small_3_2_24b_modelcard}, and Phi-4 (14B) ~\cite{abdin2024phi4technicalreport}. Their approach leverages parameter-efficient fine-tuning techniques, such as LoRA~\cite{hu2021loralowrankadaptationlarge} and adapters, allowing the models to be adapted to the task without updating all parameters. The models were trained using task-specific prompts, which were iteratively refined to improve performance across the tasks. To combine the strengths of different models, the team employed a stacking-based ensemble strategy, aggregating predictions from multiple small language models to capture complementary signals~\cite{semeval2026_task9_nycu_nlp} .

\subsubsection{Takeaways} \label{ss:task1_takeaway}
A first general trend emerging from results is the challenge of achieving consistent results on polarization detection across all languages (Table \ref{tab:subtask1}). While the best systems for each language often reach an F1-score above 0.8, performances are significantly lower for two low-resourced languages (Khmer, Burmese) and two high-resourced languages (Italian and German). 
This suggests that the intrinsic challenges in polarization detection could be caused by the lack of models' knowledge about local contexts rather than linguistic factors. Observing models that submitted results for all languages (55 out of 104), this trend seems to be confirmed. The two best performing systems ranked 10th or below on 5 languages, both struggling with Farsi, Hausa, Khmer, and Italian. 

Language-specific approaches (28 out of 104) did not perform well, though. Only two systems managed to be ranked among the top-5 in Bengali leaderboard: \textbf{CUET\-823} \cite{semeval2026_task9_cuet_823} (2) and \textbf{transformer\_1376} \cite{semeval2026_task9_transformer_1376} (5).
Finally, it is worth mentioning teams focused on specific macro-regions. It is the case of \textbf{PolAR Bears} \cite{semeval2026_task9_bears}, which submitted runs for languages spoken in Southern Asia (Bengali, Hindi, Odia, and Telugu).
Results achieved by this team were considerably subpar, though, as they consistently ranked below 10. This once again demonstrates that handling cultural variation in the computational understanding of polarization is still a major challenge in NLP research.     

\subsection{Subtask 2: \polartype}

\subsubsection{Best Performing Systems}
\begin{table}[]
\centering
\scriptsize
\setlength{\tabcolsep}{2.5pt}
\renewcommand{\arraystretch}{0.9}
\resizebox{\columnwidth}{!}{%
\begin{tabular}{lll|lll}
\toprule
\multicolumn{1}{c}{\textbf{Lang}} & \multicolumn{1}{c}{\textbf{Team}} & \multicolumn{1}{c}{\textbf{Score}} & \multicolumn{1}{c}{\textbf{Lang}} & \multicolumn{1}{c}{\textbf{Team}} & \multicolumn{1}{c}{\textbf{Score}} \\
\midrule
\multirow{4}{*}{amh} & \cellcolor{blue1}PolaFusion & \cellcolor{blue1}0.670 & \multirow{4}{*}{nep} & \cellcolor{blue3}NYCU-NLP & \cellcolor{blue3}0.810 \\
 & \cellcolor{blue1}SMASH & \cellcolor{blue1}0.650 &  & \cellcolor{blue3}Lingo Research Group & \cellcolor{blue3}0.805 \\
 & \cellcolor{blue1}YEZE & \cellcolor{blue1}0.649 &  & \cellcolor{blue3}mdok-style & \cellcolor{blue3}0.803 \\
 & \cellcolor{red1}\textit{baseline} & \cellcolor{red1}0.471 &  & \cellcolor{blue1}\textit{baseline} & \cellcolor{blue1}0.664 \\
\midrule
\multirow{4}{*}{arb} & \cellcolor{blue1}NYCU-NLP & \cellcolor{blue1}0.670 & \multirow{4}{*}{ori} & \cellcolor{blue1}UTokyo Tsuruoka Lab & \cellcolor{blue1}0.603 \\
 & \cellcolor{blue1}UTokyo Tsuruoka Lab & \cellcolor{blue1}0.668 &  & \cellcolor{red1}AIvengers & \cellcolor{red1}0.594 \\
 & \cellcolor{blue1}SMASH & \cellcolor{blue1}0.658 &  & \cellcolor{red1}NYCU-NLP & \cellcolor{red1}0.578 \\
 & \cellcolor{red1}\textit{baseline} & \cellcolor{red1}0.559 &  & \cellcolor{red1}\textit{baseline} & \cellcolor{red1}0.423 \\
\midrule
\multirow{4}{*}{ben} & \cellcolor{red1}Lingo Research Group & \cellcolor{red1}0.422 & \multirow{4}{*}{pol} & \cellcolor{blue1}UTokyo Tsuruoka Lab & \cellcolor{blue1}0.650 \\
 & \cellcolor{red1}NYCU-NLP & \cellcolor{red1}0.401 &  & \cellcolor{blue1}NYCU-NLP & \cellcolor{blue1}0.640 \\
 & \cellcolor{red2}SMASH & \cellcolor{red2}0.378 &  & \cellcolor{blue1}Lingo Research Group & \cellcolor{blue1}0.625 \\
 & \cellcolor{red2}\textit{baseline} & \cellcolor{red2}0.268 &  & \cellcolor{red1}\textit{baseline} & \cellcolor{red1}0.416 \\
\midrule
\multirow{4}{*}{deu} & \cellcolor{blue1}UTokyo Tsuruoka Lab & \cellcolor{blue1}0.620 & \multirow{4}{*}{rus} & \cellcolor{blue1}NYCU-NLP & \cellcolor{blue1}0.630 \\
 & \cellcolor{blue1}NYCU-NLP & \cellcolor{blue1}0.616 &  & \cellcolor{blue1}SMASH & \cellcolor{blue1}0.619 \\
 & \cellcolor{red1}Lingo Research Group & \cellcolor{red1}0.599 &  & \cellcolor{blue1}UTokyo Tsuruoka Lab & \cellcolor{blue1}0.617 \\
 & \cellcolor{red1}\textit{baseline} & \cellcolor{red1}0.533 &  & \cellcolor{red1}\textit{baseline} & \cellcolor{red1}0.409 \\
\midrule
\multirow{4}{*}{eng} & \cellcolor{red1}UTokyo Tsuruoka Lab & \cellcolor{red1}0.532 & \multirow{4}{*}{spa} & \cellcolor{blue1}NYCU-NLP & \cellcolor{blue1}0.681 \\
 & \cellcolor{red1}Stochastic Gradient Descenders & \cellcolor{red1}0.516 &  & \cellcolor{blue1}UTokyo Tsuruoka Lab & \cellcolor{blue1}0.674 \\
 & \cellcolor{red1}NYCU-NLP & \cellcolor{red1}0.514 &  & \cellcolor{blue1}SMASH & \cellcolor{blue1}0.673 \\
 & \cellcolor{red2}\textit{baseline} & \cellcolor{red2}0.347 &  & \cellcolor{red1}\textit{baseline} & \cellcolor{red1}0.593 \\
\midrule
\multirow{4}{*}{fas} & \cellcolor{blue1}SMASH & \cellcolor{blue1}0.644 & \multirow{4}{*}{swa} & \cellcolor{red1}SMASH & \cellcolor{red1}0.569 \\
 & \cellcolor{blue1}MSqrd & \cellcolor{blue1}0.609 &  & \cellcolor{red1}UTokyo Tsuruoka Lab & \cellcolor{red1}0.540 \\
 & \cellcolor{blue1}PolaFusion & \cellcolor{blue1}0.605 &  & \cellcolor{red1}NYCU-NLP & \cellcolor{red1}0.522 \\
 & \cellcolor{red1}\textit{baseline} & \cellcolor{red1}0.525 &  & \cellcolor{red1}\textit{baseline} & \cellcolor{red1}0.402 \\
\midrule
\multirow{4}{*}{hau} & \cellcolor{red1}NYCU-NLP & \cellcolor{red1}0.480 & \multirow{4}{*}{tel} & \cellcolor{red1}Sagarmatha & \cellcolor{red1}0.465 \\
 & \cellcolor{red1}SMASH & \cellcolor{red1}0.454 &  & \cellcolor{red1}SMASH & \cellcolor{red1}0.458 \\
 & \cellcolor{red1}Sagarmatha & \cellcolor{red1}0.427 &  & \cellcolor{red1}PolaFusion & \cellcolor{red1}0.446 \\
 & \cellcolor{red2}\textit{baseline} & \cellcolor{red2}0.216 &  & \cellcolor{red1}\textit{baseline} & \cellcolor{red1}0.426 \\
\midrule
\multirow{4}{*}{hin} & \cellcolor{blue3}SMASH & \cellcolor{blue3}0.807 & \multirow{4}{*}{tur} & \cellcolor{blue1}UTokyo Tsuruoka Lab & \cellcolor{blue1}0.652 \\
 & \cellcolor{blue3}NYCU-NLP & \cellcolor{blue3}0.801 &  & \cellcolor{blue1}NYCU-NLP & \cellcolor{blue1}0.646 \\
 & \cellcolor{blue2}YNU-HPCC & \cellcolor{blue2}0.793 &  & \cellcolor{blue1}Lingo Research Group & \cellcolor{blue1}0.624 \\
 & \cellcolor{blue2}\textit{baseline} & \cellcolor{blue2}0.700 &  & \cellcolor{red1}\textit{baseline} & \cellcolor{red1}0.484 \\
\midrule
\multirow{4}{*}{ita} & \cellcolor{red1}UTokyo Tsuruoka Lab & \cellcolor{red1}0.551 & \multirow{4}{*}{urd} & \cellcolor{blue2}Lingo Research Group & \cellcolor{blue2}0.798 \\
 & \cellcolor{red1}ShefFriday & \cellcolor{red1}0.538 &  & \cellcolor{blue2}SMASH & \cellcolor{blue2}0.790 \\
 & \cellcolor{red1}CYUT & \cellcolor{red1}0.484 &  & \cellcolor{blue2}NYCU-NLP & \cellcolor{blue2}0.789 \\
 & \cellcolor{red2}\textit{baseline} & \cellcolor{red2}0.261 &  & \cellcolor{blue2}\textit{baseline} & \cellcolor{blue2}0.739 \\
\midrule
\multirow{4}{*}{khm} & \cellcolor{blue2}UTokyo Tsuruoka Lab & \cellcolor{blue2}0.705 & \multirow{4}{*}{zho} & \cellcolor{blue3}NYCU-NLP & \cellcolor{blue3}0.844 \\
 & \cellcolor{blue2}SMASH & \cellcolor{blue2}0.702 &  & \cellcolor{blue3}UTokyo Tsuruoka Lab & \cellcolor{blue3}0.835 \\
 & \cellcolor{blue1}PolaFusion & \cellcolor{blue1}0.699 &  & \cellcolor{blue3}Lingo Research Group & \cellcolor{blue3}0.825 \\
 & \cellcolor{red1}\textit{baseline} & \cellcolor{red1}0.586 &  & \cellcolor{blue1}\textit{baseline} & \cellcolor{blue1}0.631 \\
\midrule
\multirow{4}{*}{mya} & \cellcolor{blue2}SMASH & \cellcolor{blue2}0.736 & \multicolumn{1}{l}{} &  & \multicolumn{1}{l}{} \\
 & \cellcolor{blue2}UTokyo Tsuruoka Lab & \cellcolor{blue2}0.708 & \multicolumn{1}{l}{} &  & \multicolumn{1}{l}{} \\
 & \cellcolor{blue2}PolarMind & \cellcolor{blue2}0.702 & \multicolumn{1}{l}{} &  & \multicolumn{1}{l}{} \\
 & \cellcolor{red1}\textit{baseline} & \cellcolor{red1}0.551 & \multicolumn{1}{l}{} &  & \multicolumn{1}{l}{}\\
\cmidrule{1-3}
\end{tabular}
}
\caption{Top three performing systems for each language in subtask~2 evaluated using macro-F1 score.}
\label{tab:subtask2}
\end{table}

\textbf{Team UTokyo Tsuruoka Lab} managed to score first place in Subtask 2 in 7 out of the 22 languages, making them the best-scoring team. They used the same models and fine-tuning tools as their efforts in Subtask 1, they performed key modifications to account for the multilabel setup. They used JSON finetuning as an auto-regressive baseline, where they instructed the model to generate JSON objects with a binary decision for each label, to later use it during training and inference with cross-entropy loss and greedy rule-based approach, respectively. Finally, they adapted SALSA for a multi-label classification \cite{semeval2026_task9_utokyo_tsuruoka_lab}.

\textbf{Team NYCU-NLP} found difficulty in the ``Other" category, and therefore implemented a heuristic based on the prediction made in Subtask 1, using it as an auxiliary signal during inference. With this modification to their initial approach, the team managed to land in first place for 6 of the 22 languages, close second to the best performing team \cite{semeval2026_task9_nycu_nlp}.

\subsubsection{Takeaways}
Results of Subtask 2 (Table \ref{tab:subtask2}) are significantly lower than Subtask 1, with 7 languages in which the highest F-score was below 0.6 and only 3 languages with a score above 0.8. No specific trends about language families and macro-regions emerge, though. 

Similarly to what has been observed in Section \ref{ss:task1_takeaway}, the highest ranked models exhibit a drop in performance related to specific languages, even if they are not the same from the previous task. E.g., team \textbf{UTokyo Tsuruoka Lab}, which ranked 25th in Subtask 1 for Italian, ranked first in Subtask 2. 

Additional insights from the task emerge from model performances across different languages and topics. Table \ref{tab:topic_bias} reports the percentage of polarization types correctly predicted by all the models that participated in the tasks (true positives). As it can be observed, a strong cultural variation seems to emerge across languages. E.g., the percentage of correct prediction of Gender/Sexual polarity types is 0.239 for Amharic and 0.825 for Chinese. Such oscillation is also present in languages from the same macro-region. For instance, only 0.365 Religious polarity types are correctly identified in Telugu; 0.905 in Hindi. Therefore, the generalization of polarity types across different languages and local contexts remains an open issue for the NLP research community.

\subsection{Subtask 3: \polarmanifest}

\begin{table}[h]
\scriptsize
\centering
\setlength{\tabcolsep}{2.5pt}
\renewcommand{\arraystretch}{0.9}
\resizebox{\columnwidth}{!}{%
\begin{tabular}{@{}lll|lll@{}}
\toprule
\multicolumn{1}{c}{\textbf{Lang}} & \multicolumn{1}{c}{\textbf{Team}} & \multicolumn{1}{c}{\textbf{Score}} & \multicolumn{1}{c}{\textbf{Lang}} & \multicolumn{1}{c}{\textbf{Team}} & \multicolumn{1}{c}{\textbf{Score}} \\
\toprule
\multirow{4}{*}{amh} & \cellcolor{red1}SMASH & \cellcolor{red1}0.579 & \multirow{4}{*}{nep} & \cellcolor{blue2}NYCU-NLP & \cellcolor{blue2}0.713 \\
 & \cellcolor{red1}NYCU-NLP & \cellcolor{red1}0.559 &  & \cellcolor{blue2}SMASH & \cellcolor{blue2}0.712 \\
 & \cellcolor{red1}AIvengers & \cellcolor{red1}0.554 &  & \cellcolor{blue1}Lingo Research Group & \cellcolor{blue1}0.669 \\
 & \cellcolor{red1}\textit{baseline} & \cellcolor{red1}0.512 &  & \cellcolor{blue1}\textit{baseline} & \cellcolor{blue1}0.602 \\
\midrule
\multirow{4}{*}{arb} & \cellcolor{blue1}NYCU-NLP & \cellcolor{blue1}0.646 & \multirow{4}{*}{ori} & \cellcolor{red2}SMASH & \cellcolor{red2}0.330 \\
 & \cellcolor{blue1}SMASH & \cellcolor{blue1}0.641 &  & \cellcolor{red2}Ping An & \cellcolor{red2}0.328 \\
 & \cellcolor{blue1}YEZE & \cellcolor{blue1}0.610 &  & \cellcolor{red2}NYCU-NLP & \cellcolor{red2}0.297 \\
 & \cellcolor{red1}\textit{baseline} & \cellcolor{red1}0.568 &  & \cellcolor{red2}\textit{baseline} & \cellcolor{red2}0.240 \\
\midrule
\multirow{4}{*}{ben} & \cellcolor{red2}SMASH & \cellcolor{red2}0.281 & \multirow{4}{*}{pan} & \cellcolor{red1}NYCU-NLP & \cellcolor{red1}0.544 \\
 & \cellcolor{red2}Ping An & \cellcolor{red2}0.255 &  & \cellcolor{red1}SMASH & \cellcolor{red1}0.541 \\
 & \cellcolor{red2}PolaFusion & \cellcolor{red2}0.249 &  & \cellcolor{red1}AIvengers & \cellcolor{red1}0.529 \\
 & \cellcolor{red2}\textit{baseline} & \cellcolor{red2}0.258 &  & \cellcolor{red1}\textit{baseline} & \cellcolor{red1}0.484 \\
\midrule
\multirow{4}{*}{deu} & \cellcolor{red1}NYCU-NLP & \cellcolor{red1}0.518 & \multirow{4}{*}{spa} & \cellcolor{red1}SMASH & \cellcolor{red1}0.541 \\
 & \cellcolor{red1}ShefFriday & \cellcolor{red1}0.515 &  & \cellcolor{red1}NYCU-NLP & \cellcolor{red1}0.520 \\
 & \cellcolor{red1}SMASH & \cellcolor{red1}0.513 &  & \cellcolor{red1}PolaFusion & \cellcolor{red1}0.507 \\
 & \cellcolor{red1}\textit{baseline} & \cellcolor{red1}0.471 &  & \cellcolor{red1}\textit{baseline} & \cellcolor{red1}0.480 \\
\midrule
\multirow{4}{*}{eng} & \cellcolor{red1}Sagarmatha & \cellcolor{red1}0.511 & \multirow{4}{*}{swa} & \cellcolor{red1}SMASH & \cellcolor{red1}0.584 \\
 & \cellcolor{red1}Ping An & \cellcolor{red1}0.507 &  & \cellcolor{red1}AIvengers & \cellcolor{red1}0.565 \\
 & \cellcolor{red1}SMASH & \cellcolor{red1}0.507 &  & \cellcolor{red1}OZemi & \cellcolor{red1}0.562 \\
 & \cellcolor{red1}\textit{baseline} & \cellcolor{red1}0.466 &  & \cellcolor{red1}\textit{baseline} & \cellcolor{red1}0.565 \\
\midrule
\multirow{4}{*}{fas} & \cellcolor{red1}SMASH & \cellcolor{red1}0.493 & \multirow{4}{*}{tel} & \cellcolor{red1}SMASH & \cellcolor{red1}0.445 \\
 & \cellcolor{red1}OZemi & \cellcolor{red1}0.476 &  & \cellcolor{red1}PolaFusion & \cellcolor{red1}0.429 \\
 & \cellcolor{red1}Sagarmatha & \cellcolor{red1}0.461 &  & \cellcolor{red1}Sagarmatha & \cellcolor{red1}0.424 \\
 & \cellcolor{red2}\textit{baseline} & \cellcolor{red2}0.395 &  & \cellcolor{red2}\textit{baseline} & \cellcolor{red2}0.392 \\
\midrule
\multirow{4}{*}{hau} & \cellcolor{red2}Sagarmatha & \cellcolor{red2}0.207 & \multirow{4}{*}{tur} & \cellcolor{red1}NYCU-NLP & \cellcolor{red1}0.538 \\
 & \cellcolor{red2}OZemi & \cellcolor{red2}0.206 &  & \cellcolor{red1}Ping An & \cellcolor{red1}0.537 \\
 & \cellcolor{red2}PolaFusion & \cellcolor{red2}0.204 &  & \cellcolor{red1}PolaFusion & \cellcolor{red1}0.515 \\
 & \cellcolor{red2}\textit{baseline} & \cellcolor{red2}0.206 &  & \cellcolor{red1}\textit{baseline} & \cellcolor{red1}0.449 \\
\midrule
\multirow{4}{*}{hin} & \cellcolor{blue2}SMASH & \cellcolor{blue2}0.771 & \multirow{4}{*}{urd} & \cellcolor{blue3}NYCU-NLP & \cellcolor{blue3}0.821 \\
 & \cellcolor{blue2}NYCU-NLP & \cellcolor{blue2}0.770 &  & \cellcolor{blue3}SMASH & \cellcolor{blue3}0.821 \\
 & \cellcolor{blue2}PolaFusion & \cellcolor{blue2}0.759 &  & \cellcolor{blue3}YEZE & \cellcolor{blue3}0.815 \\
 & \cellcolor{blue2}\textit{baseline} & \cellcolor{blue2}0.701 &  & \cellcolor{blue2}\textit{baseline} & \cellcolor{blue2}0.771 \\
\midrule
\multirow{4}{*}{khm} & \cellcolor{red1}SMASH & \cellcolor{red1}0.437 & \multirow{4}{*}{zho} & \cellcolor{blue2}NYCU-NLP & \cellcolor{blue2}0.719 \\
 & \cellcolor{red1}PolaFusion & \cellcolor{red1}0.400 &  & \cellcolor{blue2}CYUT & \cellcolor{blue2}0.700 \\
 & \cellcolor{red2}AIvengers & \cellcolor{red2}0.377 &  & \cellcolor{blue1}SMASH & \cellcolor{blue1}0.677 \\
 & \cellcolor{red2}\textit{baseline} & \cellcolor{red2}0.343 &  & \cellcolor{red1}\textit{baseline} & \cellcolor{red1}0.461\\
\bottomrule
\end{tabular}
}
\caption{Top three performing systems for each language in subtask~3 evaluated using macro-F1 score.}
\label{tab:subtask3}
\end{table}

\subsubsection{Best Performing Systems}
\textbf{Team SMASH} achieved strong performance in the competition, ranking first in 9 out of 18 languages. Their system relies on full model fine-tuning and uses 5-fold cross-validation with three random seeds for each language. Logits are averaged across seeds and folds to obtain out-of-fold predictions, which are then used to tune per-language ensemble weights and label-specific thresholds that maximize macro-F1. For final predictions, the model is retrained on all training data, logits are averaged across seeds, and the optimized weights and thresholds are applied to generate the final labels \cite{semeval2026_task9_smash}.

\textbf{Team NYCU-NLP} changed little in their approach from their Subtask~2. However, they still manage to come in first place for multiple languages, a total of 7 from the 18 languages pool for this Subtask \cite{semeval2026_task9_nycu_nlp}. 

\subsubsection{Takeaways}
As with the previous Subtask, a decrease in performance is noticeable but in a more dramatic tone. Table \ref{tab:subtask3} shows the best performing systems, and for only one language, Urdu, the score was above 0.8. Furthermore, a score above 0.7 was achieved by only three more languages, and for five languages the score was below 0.5. A similar trend can be seen in Table~\ref{tab:topic_bias}, where correct labeling did not improve much from the previous subtask. 

It is worth noting that the best-performing languages are from the region of Southern Asia: Hindi, Nepali, and Urdu. In these languages, the SMASH \citep{semeval2026_task9_smash} and NYCU-NLP \citep{semeval2026_task9_nycu_nlp} teams either tied or are very close in their score. It is assumed that their approaches perform specifically well for these languages, as their scores in other languages fall drastically behind.

\section{Discussion}

\subsection{Popular Methods}

The most common methods include ensemble prediction, fine-tuning, threshold tuning for language or class labels, and data augmentation. 

\paragraph{Model Families} The Qwen family~\cite{bai2023qwentechnicalreport} is the most frequently used model (31\%), followed by the LLaMA family~\cite{touvron2023llamaopenefficientfoundation} (20\%)~\, and the Gemma/Gemini~\cite{gemmateam2025gemma3technicalreport} family (19\%). Several teams also employed GPT~\cite{openai2024gpt4technicalreport}, Mistral~\cite{mistral_small_3_2_24b_modelcard}, and BERT-based encoder models (each 7\%), while Deepseek~\cite{deepseekai2025deepseekv3technicalreport}, Phi~\cite{abdin2024phi4technicalreport}, GLM~\cite{5team2025glm45agenticreasoningcoding}, and Nemotron~\cite{nvidia2024nemotron4340btechnicalreport} are used in a only small number of systems.

\paragraph{Ensemble models} Model ensembling is one of the commonly used techniques. Methods include ensembling multiple transformer models, combining transformer encoders with LLMs, or integrating models from different architectural families. Teams adopted various strategies to determine ensemble weights, including learning weights from out-of-folder logits, soft-voting ensembles, and weighted or average fusion. 

\paragraph{Fine tuning} Approximately 39\% of teams reported applying fine-tuning, while half of them employ parameter-efficient fine-tuning (PEFT) such as LoRA~\cite{hu2021loralowrankadaptationlarge}.

\paragraph{Loss optimization} Because the multi-label subtasks involved heavily imbalanced distributions, standard cross-entropy was frequently replaced with more robust loss optimization techniques. Popular alternatives included Asymmetric Loss (ASL), Weighted Binary Cross-Entropy, and Focal Loss.

\paragraph{Data augmentation} Approximately 38\% of teams reported using data augmentation to mitigate class or language imbalance. Common techniques include back-translation, cross-lingual translation, extending instances with generated explanation, paraphrasing, hard-negative generation, or easy data augmentation like lowercasing, uppercasing, shuffling words, or replacing them with synonyms.

\paragraph{Per-label and per-language threshold calibration} Most systems report using per-label or per-language threshold tuning to address underrepresented label or language distribution, often improving performance in imbalanced settings.

\subsection{Best performing Systems}

Based on the overall ranking statistics (See Table~\ref{tab:medal_by_task}) across languages and subtasks, we highlight three teams that demonstrated particularly strong performance in the shared task: UTokyo Tsuruoka Lab~\cite{semeval2026_task9_utokyo_tsuruoka_lab},  NYCU-NLP~\cite{semeval2026_task9_nycu_nlp}, and SMASH~\cite{semeval2026_task9_smash}. UTokyo Tsuruoka Lab achieved the most first-place rankings across Subtask~1 and Subtask~2, indicating strong peak performance. Team NYCU-NLP obtained 38 top-3 placements across all subtasks, the highest among participating teams. Team SMASH also achieved competitive results, ranking first in Subtask~ 3 and obtaining 36 top-3 placements across the evaluation.

The strategies behind these strong results differ across teams. UTokyo fine-tuned Gemma-3-12B-IT and Gemma-3-27B-IT-bnb-4bit~\cite{gemmateam2025gemma3technicalreport} using LoRA~\cite{hu2021loralowrankadaptationlarge}. They attribute their performance to a single-forward-pass inference paradigm rather than JSON-format inference. NYCU-NLP employed a stacking-based ensemble strategy using three LLMs: Gemma-3 (27B)~\cite{gemmateam2025gemma3technicalreport}, Mistral Small 3.2 (24B)~\cite{mistral_small_3_2_24b_modelcard}, and Phi-4 (14B)~\cite{abdin2024phi4technicalreport}, and also introduced a heuristic method for predicting the ``other'' category in Subtask~2. SMASH adopted an ensemble approach that combines monolingual and multilingual encoder-based transformers, including mDeBERTa~\cite{he2023debertav}, XLM-R~\cite{conneau-etal-2020-unsupervised}, and mBERT~\cite{devlin2019bert}. In addition, they attribute their performance to out-of-fold ensemble weight tuning and per-class threshold calibration.

\section{Conclusion}
This shared task has attracted over 1,000 participants, with 69 system description papers submitted, making it the most popular task among all 12 SemEval/-2026 tasks. While most participants adopted commonly used strategies, such as data augmentation, fine-tuning, ensemble models, and per-class or per-language threshold calibration, the top-performance teams employed different approaches. No single method dominates, and strong performance can be achieved through multiple strategies.

We created a successful and challenging experience for the computational linguistics community. Through an engaging team and willing organizers, the task was the most involved task for the SemEval-2026. Bringing forward the pressing issue of polarization that occurs in many cultures and languages, through multiple events, many interesting approaches emerged, and this communal effort has fostered research opportunities and collaboration. As a byproduct, the remarkable dataset has been created and made public to help future research on polarization.

\clearpage
\section*{Limitations}

We consider our task an important step towards multilingual, multicultural, and multi-event polarization analysis, but several limitations remain. Particularly on the quality assurance of the labeling, as some of the languages utilized crowdsource, and it often comes with a disadvantage in the full grasp and understanding of the task at hand from the annotators. Quality assessments, like pilots and control questions, were placed to help with this, but some inconsistencies may linger.

For some of the languages in our task, the available data was limited, which could limit the generalizability of the resulting models or may have resulted in abstention from participation. Future efforts or iterations should strive for wider size and diversity, and possibly explore language- or region-specific data points to platform underrepresented communities, as well as NLP researchers.

\subsection*{Ethics Statement}

Throughout the task, we strive to have an open channel of communication and support for the participants, as well as warn them of the potential triggering content they attempted to classify. 

The annotators of the dataset were also warned and given resources to handle the stress the annotation may have brought forward. They were fairly paid, in accordance with their local laws and, where valid, the standard of the crowdsourcing tools. All the annotators were either native speakers of their respective languages, hoping to capture their unique insight into polarization for their culture and language. 

Polarization is a sensitive and, up to a degree, subjective topic. The resulting dataset was made public, and the data was anonymized, which therefore begs for its further usage to be in a responsible and ethical manner.

\subsection{Acknowledgments}
We thank the SemEval-2026 organizers for the opportunity to take our tasks into the international research community, and the participants for taking part in it in a meaningful and eager way.

We thank all annotators who participated in annotating each language for their contributions and efforts.

The University of Hamburg (UHH) team acknowledges the grant from the Google Award for Inclusion Research Program, which supports AI-MAP\footnote{\url{https://www.hcds.uni-hamburg.de/en/research/ai-map.html}} project that results in the extension of POLAR project.

Tanmoy Chakraborty acknowledges the financial support of Anusandhan National Research Foundation (CRG/2023/001351).

Ö. Alacam received funding through the project SAIL: SustAInable Life-cycle of Intelligent Socio-Technical Systems (Grant ID NW21059A), funded by the Ministry of Culture and Science of the State of North Rhine-Westphalia (Germany).

Shamsuddeen Hassan Muhammad acknowledges the support of Google DeepMind, whose funding made this work possible.

Cengiz Acartürk thank Jagiellonian University Strategic Programme Excellence Initiative ID.UJ for providing partial financial support, and Anna Maria Wilkosz and Jakub Romanowski for the annotations.

Usman Naseem acknowledges the support of the DAAD Research Fellowship, which supported the initiation of this work.

The work of Elena Tutubalina was supported within the framework of the HSE University Basic Research Program, and the computational resources of HSE University’s HPC facilities are acknowledged.

\bibliography{custom,participant_papers}

\clearpage

\appendix

\section{Label distribution}
\begin{table*}[!t]
\centering
\scriptsize
\setlength{\tabcolsep}{3.2pt}
\renewcommand{\arraystretch}{1}
\resizebox{\textwidth}{!}{%
\begin{tabular}{l| r|r| rrrrr| rrrrrr}
\toprule
\multirow{2}{*}{\textbf{Lang.}} &
\multirow{2}{*}{\textbf{Total}} &
\multicolumn{1}{c}{\textbf{Subtask 1}} &
\multicolumn{5}{c}{\textbf{Subtask 2}} &
\multicolumn{6}{c}{\textbf{Subtask 3}} \\
\cmidrule(lr){3-3}
\cmidrule(lr){4-8}
\cmidrule(lr){9-14}
& &
\textbf{\shortstack{Polarized \\ (\%)}} &
\textbf{Political} &
\textbf{\shortstack{Racial / \\ Ethnic}} &
\textbf{\shortstack{Religious \\ Polarization}} &
\textbf{\shortstack{Gender / \\ Sexual}} &
\textbf{Other} &
\textbf{\shortstack{Stereo- \\ type}} &
\textbf{\shortstack{Vilifi- \\ cation}} &
\textbf{\shortstack{Dehuman- \\ ization}} &
\textbf{\shortstack{Extreme \\ Language}} &
\textbf{\shortstack{Lack of \\ Empathy}} &
\textbf{\shortstack{Invalid- \\ ation}} \\
\midrule
eng & 4,834 & 37\% & 36\% & 9\% & 3\% & 2\% & 4\% & 15\% & 26\% & 12\% & 24\% & 11\% & 18\% \\
deu & 4,771 & 48\% & 41\% & 19\% & 11\% & 6\% & 14\% & 36\% & 30\% & 15\% & 22\% & 27\% & 16\% \\
urd & 5,346 & 69\% & 67\% & 54\% & 55\% & 51\% & 51\% & 62\% & 65\% & 56\% & 62\% & 56\% & 57\% \\
hin & 4,117 & 85\% & 74\% & 12\% & 59\% & 11\% & 13\% & 50\% & 65\% & 18\% & 51\% & 57\% & 66\% \\
ben & 5,000 & 43\% & 34\% & 1\% & 2\% & 1\% & 10\% & 6\% & 24\% & 11\% & 5\% & 2\% & 2\% \\
ori & 3,552 & 29\% & 21\% & 5\% & 6\% & 3\% & 4\% & 10\% & 11\% & 1\% & 13\% & 2\% & 3\% \\
pan & 2,609 & 49\% & 31\% & 6\% & 8\% & 11\% & 9\% & 16\% & 40\% & 22\% & 24\% & 12\% & 24\% \\
nep & 3,008 & 50\% & 17\% & 14\% & 8\% & 5\% & 12\% & 27\% & 31\% & 7\% & 27\% & 11\% & 15\% \\
fas & 4,943 & 74\% & 44\% & 2\% & 10\% & 6\% & 24\% & 13\% & 58\% & 4\% & 17\% & 10\% & 8\% \\
ita & 5,038 & 43\% & 8\% & 18\% & 7\% & 9\% & 4\% & - & - & - & - & - & - \\
spa & 4,958 & 50\% & 27\% & 19\% & 16\% & 13\% & 13\% & 27\% & 31\% & 9\% & 24\% & 24\% & 11\% \\
rus & 5,023 & 30\% & 14\% & 10\% & 4\% & 6\% & 2\% & - & - & - & - & - & - \\
pol & 3,587 & 42\% & 37\% & 9\% & 4\% & 5\% & 6\% & - & - & - & - & - & - \\
arb & 5,070 & 45\% & 24\% & 17\% & 8\% & 11\% & 17\% & 33\% & 37\% & 11\% & 30\% & 17\% & 8\% \\
amh & 4,999 & 75\% & 67\% & 26\% & 2\% & 1\% & 25\% & 55\% & 48\% & 13\% & 31\% & 18\% & 16\% \\
hau & 5,477 & 11\% & 5\% & 3\% & 3\% & 1\% & 0\% & 4\% & 1\% & 4\% & 3\% & 1\% & 0\% \\
zho & 6,421 & 50\% & 6\% & 23\% & 2\% & 17\% & 9\% & 30\% & 19\% & 5\% & 8\% & 8\% & 5\% \\
mya & 4,334 & 58\% & 25\% & 5\% & 3\% & 11\% & 45\% & - & - & - & - & - & - \\
khm & 9,960 & 91\% & 18\% & 1\% & 3\% & 2\% & 66\% & 68\% & 2\% & 1\% & 2\% & 11\% & 7\% \\
tel & 3,550 & 53\% & 22\% & 17\% & 9\% & 13\% & 24\% & 11\% & 22\% & 2\% & 13\% & 26\% & 23\% \\
swa & 10,487 & 50\% & 3\% & 35\% & 4\% & 2\% & 8\% & 40\% & 41\% & 13\% & 24\% & 30\% & 23\% \\
tur & 3,566 & 50\% & 44\% & 16\% & 16\% & 6\% & 5\% & 41\% & 33\% & 11\% & 44\% & 10\% & 4\% \\
\midrule
Total & 110,650 & 53\% & 28\% & 16\% & 10\% & 8\% & 19\% & 28\% & 26\% & 10\% & 18\% & 16\% & 14\%\\
\bottomrule
\end{tabular}
}
\vspace{-0.25cm}
\caption{Proportion of correct label $=1$ for each topic across languages (ISO codes).}
\label{tab:topic_bias}
\end{table*}

\section{Participants}

\begin{table*}
\centering
\scriptsize
\resizebox{\textwidth}{!}{%
\begin{tabular}{l|c|p{9cm}|l}
\toprule
\textbf{Team Name} & \textbf{Attended Tasks} & \textbf{Affiliation} & \textbf{Publication} \\
\midrule
Aaron &1 & African Institute for Mathematical Sciences &  ~\cite{semeval2026_task9_aaron} \\
\rowcolor[HTML]{EFEFEF} Aatman &1 & University of Tübingen &  ~\cite{semeval2026_task9_aatman} \\
ABARUAH &1,2,3 & Assam Don Bosco University &  ~\cite{semeval2026_task9_abaruah} \\
\rowcolor[HTML]{EFEFEF} AI4PC-Howard University & 1 & Howard University & ~\cite{semeval2026_task9_ai4pc}\\
AIvengers &1,2,3 & University of Augsburg&  ~\cite{semeval2026_task9_aivengers}\\
\rowcolor[HTML]{EFEFEF} AlphaLyrae &1,2,3 & University of Information Technology, Ho Chi Minh City; Vietnam National University& ~\cite{semeval2026_task9_alphalyrae} \\
BITS Pilani & 1,2,3 & University of Information Technology; Vietnam National University & ~\cite{semeval2026_task9_pilani}\\
\rowcolor[HTML]{EFEFEF} CoPol & 2 & Delhi Skill and Entrepreneurship University & ~\cite{semeval2026_task9_copol}\\
CUET-823 &1,2 &Chittagong University of Engineering and Technology & ~\cite{semeval2026_task9_cuet_823} \\
\rowcolor[HTML]{EFEFEF} CYUT &1,2,3 & Chaoyang University of Technology & ~\cite{semeval2026_task9_cyut} \\
DataBees &1 & Sri Sivasubramaniya Nadar College of Engineering & ~\cite{semeval2026_task9_databees} \\
\rowcolor[HTML]{EFEFEF} DeepSemantics &3 & African Institute for Mathematical Sciences (AIMS) & ~\cite{semeval2026_task9_deepsemantics} \\
DigiS-FBK &1 & Fondazione Bruno Kessler; University of Trento &  ~\cite{semeval2026_task9_digis_fbk}\\
\rowcolor[HTML]{EFEFEF} DUTH &1 & Democritus University of Thrace & ~\cite{semeval2026_task9_duth} \\
Gradient Descenders &2 & University of Information Technology; National University & ~\cite{strich2025encourageevaluatingraglocal} \\
\rowcolor[HTML]{EFEFEF} HausaNLP & 2 & ACETEL; HausaNLP; Gombe State University; Nasarawa State University &  ~\cite{semeval2026_task9_hausanlp} \\
ILA-POLAR & 2 & Universität Tübingen & ~\cite{semeval2026_task9_submission_394}\\
\rowcolor[HTML]{EFEFEF} ILab-NLP &1 & Heriot-Watt University Heriot-Watt University Heriot-Watt University &  ~\cite{semeval2026_task9_ilab_nlp}\\
INFOTEC-NLP &1 & INFOTEC; SECIHTI& ~\cite{semeval2026_task9_infotec_nlp} \\
\rowcolor[HTML]{EFEFEF} IReL\_IIT(BHU) &1,2,3 & Indian Institute of Technology (BHU) Varanasi & ~\cite{semeval2026_task9_irel_iit_bhu} \\
JAT &1 & Universität Tübingen &  ~\cite{semeval2026_task9_jat}\\
\rowcolor[HTML]{EFEFEF} Joshualee2 &1 & De Anza College & ~\cite{semeval2026_task9_joshualee2} \\
Lingo Research Group &1,2,3 & Noida Institute of Engineering and Technology; IIT Gandhinagar& ~\cite{semeval2026_task9_lingo_research_group} \\
\rowcolor[HTML]{EFEFEF} mdok-style &1,2,3 & Kempelen Institute of Intelligent Technologies; ADAPTCentre,Trinity College Dublin & ~\cite{semeval2026_task9_mdok_style} \\
MINDS &1,2 & Politecnico di Torino & ~\cite{semeval2026_task9_minds} \\
\rowcolor[HTML]{EFEFEF} MKJ &1 & University of Turin &  ~\cite{semeval2026_task9_mkj}\\
MoMo &1 & University of Delhi Delhi Skill and Entrepreneurship University & ~\cite{semeval2026_task9_momo} \\
\rowcolor[HTML]{EFEFEF} MSqrd &1,2,3 & Habib University & ~\cite{semeval2026_task9_msqrd} \\
NAMAA &1,2 & NAMAA Community; Datategy &  ~\cite{semeval2026_task9_namaa_491}\\
\rowcolor[HTML]{EFEFEF} NASIM\_Lab &2 & The University of Western Australia & ~\cite{semeval2026_task9_nasim_lab} \\
NIT-Agartala-NLP-Team &1,2 & National Institute of Technology Agartala& ~\cite{semeval2026_task9_nit_agartala_nlp_team} \\
\rowcolor[HTML]{EFEFEF} NLP-CIMAT &1 & CIMAT; SECIHTI& ~\cite{semeval2026_task9_nlp_cimat}\\
NYCU-NLP &1,2,3 & National Yang Ming Chiao Tung University & ~\cite{semeval2026_task9_nycu_nlp} \\
\rowcolor[HTML]{EFEFEF} OZemi &1,2,3 & Waseda University &  ~\cite{semeval2026_task9_ozemi}\\
pfr821& 1 & Télécom Paris; Airbus Defence \& Space & ~\cite{semeval2026_task9_pfr812}\\
\rowcolor[HTML]{EFEFEF} Phatthachdau &1 & VNUHCM-University of Information Technology & ~\cite{semeval2026_task9_phatthachdau} \\
PolaFusion &1,2,3 & Delhi Skill and Entrepreneurship University (DSEU) & ~\cite{semeval2026_task9_polafusion} \\
\rowcolor[HTML]{EFEFEF} PolAR Bears &1,2,3 & Oogwai Analytics; Banaras Hindu University & ~\cite{semeval2026_task9_bears} \\
PolarizedTeam &1,2 & “Alexandru Ioan Cuza” University of Iasi; Romanian Academy- Iasi Branch & ~\cite{semeval2026_task9_polarizedteam} \\
\rowcolor[HTML]{EFEFEF} PolarMind &1,2 & Indian Institute of Technology& ~\cite{semeval2026_task9_polarmind_446} \\
PolDeck &1,2 & University of Augsburg & ~\cite{semeval2026_task9_poldeck} \\
\rowcolor[HTML]{EFEFEF} PSK &1 & Independent Researcher & ~\cite{semeval2026_task9_psk} \\
REGLAT &1 & Benha University; College of Engineering; University of Al-Kharj; SUTech & ~\cite{semeval2026_task9_reglat} \\
\rowcolor[HTML]{EFEFEF} Sagarmatha &1,2,3 & IIMSCollege; PCPSCollege& ~\cite{semeval2026_task9_sagarmatha} \\
Seals-NLP &1 & Auburn University& ~\cite{semeval2026_task9_seals_nlp} \\
\rowcolor[HTML]{EFEFEF} Semantic Vectors &1 & IIT Dharwad & ~\cite{semeval2026_task9_vectors} \\
ServSocIA &1 & Universidad de la República; Aplicadas y en Sistemas &~\cite{semeval2026_task9_servsocia}\\
\rowcolor[HTML]{EFEFEF} ShefFriday &1,2,3 & The University of Sheffield & ~\cite{semeval2026_task9_sheffriday} \\
SMASH &1,2,3 & University of Edinburgh & ~\cite{semeval2026_task9_smash} \\
\rowcolor[HTML]{EFEFEF} StanceLab &1 & University of Iasi & ~\cite{semeval2026_task9_stancelab} \\
Stochastic Gradient Descenders &2 & University of Information Technology; Vietnam National University& ~\cite{semeval2026_task9_descenders_144} \\
\rowcolor[HTML]{EFEFEF} Taien &1 & BGC Trust University; University of Chittagong & ~\cite{semeval2026_task9_taien} \\
The Argonauts &1,2 & Chittagong University of Engineering and Technology &  ~\cite{semeval2026_task9_argonauts}\\
\rowcolor[HTML]{EFEFEF} The Counterfactuals &1,2,3 & University of Colorado & ~\cite{semeval2026_task9_counterfactuals} \\
Tralaleros &1 & Kiel University; University of Hamburg &  ~\cite{semeval2026_task9_tralaleros}\\
\rowcolor[HTML]{EFEFEF} transformer\_1376 &1 & Chittagong University of Engineering \& Technology & ~\cite{semeval2026_task9_transformer_1376} \\
UIT-Polar &1 & University of Information Technology; Vietnam National University & ~\cite{semeval2026_task9_uit_polar} \\
\rowcolor[HTML]{EFEFEF} UMUSP &1,2,3 & University of Minho & ~\cite{semeval2026_task9_umusp} \\
UNED &1 & NLP \& IR group at UNED & ~\cite{semeval2026_task9_uned} \\
\rowcolor[HTML]{EFEFEF} UPR &1,2,3 & Sejong University & ~\cite{semeval2026_task9_upr,semeval2026_task9_upr_369,semeval2026_task9_upr_370} \\
UTokyo Tsuruoka Lab &1,2 & The University of Tokyo & ~\cite{semeval2026_task9_utokyo_tsuruoka_lab} \\
\rowcolor[HTML]{EFEFEF} VGU-M.Tech-AI &2 & Vivekananda Global University Jaipur & ~\cite{semeval2026_task9_vgu_m_tech_ai} \\
wangkongqiang &1,2,3 & Yunnan University & ~\cite{semeval2026_task9_wangkongqiang} \\
\rowcolor[HTML]{EFEFEF} YEZE &1,2,3 & University of Tübingen &  ~\cite{semeval2026_task9_yeze}\\
YNU-HPCC &2 & Yunnan University & ~\cite{semeval2026_task9_ynu_hpcc} \\
\rowcolor[HTML]{EFEFEF} zhangpeng &1,2,3 & Yunnan University; Yunnan Province Smart Tourism Engineering Research Center& ~\cite{semeval2026_task9_zhangpeng}\\
\bottomrule
\end{tabular}
}
\caption{Overview of participating teams, their attended subtasks, affiliations, and publications.}
\label{tab:participants}
\end{table*}

\end{document}